%% file: root.tex
\documentclass[10pt,twocolumn,letterpaper]{article}

\usepackage{wacv}              

\usepackage{cite}
\usepackage{amsmath,graphicx}
\usepackage[pagebackref,breaklinks,colorlinks]{hyperref}
\usepackage{soul,color}
\usepackage{caption}
\usepackage{diagbox}
\usepackage{amssymb}
\usepackage{textcomp}
\usepackage{multirow}
\usepackage{booktabs}
\usepackage{makecell}
\usepackage{bbding}
\usepackage{pifont}
\usepackage{xcolor}
\usepackage{colortbl}
\usepackage{hhline}

\def\wacvPaperID{*****} 
\def\confName{WACV}
\def\confYear{2025}

\title{\LARGE \bf
Reflective Teacher: Semi-Supervised Multimodal 3D Object Detection in Bird's-Eye-View via Uncertainty Measure
}

\author{\parbox{0.85\linewidth}{\centering Saheli Hazra\textsuperscript{1}, Sudip Das\textsuperscript{2}, Rohit Choudhary\textsuperscript{3}, Arindam Das\textsuperscript{2,4}, Ganesh Sistu\textsuperscript{4,5},\\ Ciar\'{a}n Eising\textsuperscript{4}, and  Ujjwal Bhattacharya\textsuperscript{1}}\\
$^{1}$CVPR Unit, Indian Statistical Institute, Kolkata, 
$^{2}$anSWer, Valeo India, 
$^{3}$IIT Madras, India, \\
$^{4}$University of Limerick, Ireland, and 
$^{5}$Valeo Vision Systems, Ireland\\
$^{1}${\tt\small {\{saheli2806\_r, ujjwal\}}@{isical.ac.in, }}
\tt\small firstname.lastname@{\{valeo.com, ul.ie\}},\\
$^{3}${\tt\small rohitc.manth@{gmail.com}}\\
}

\ifx01 
\newcommand{\AD}[1]{\textcolor{blue}{[Arindam: #1]}}
\else 
\newcommand{\AD}[1]{\textcolor{blue}{}}
\fi

\begin{document}

\maketitle


\input{include/abstract.tex}

\input{include/introduction.tex}

\input{include/related_work.tex}

\input{include/architecture.tex}

\input{include/results.tex}

\input{include/conclusion.tex}



\bibliographystyle{ieeetr}
\bibliography{IEEEexample}

\end{document}

%% file: include/abstract.tex
\begin{abstract}

Applying pseudo labeling techniques has been found to be advantageous in semi-supervised 3D object detection (SSOD) in Bird’s-Eye-View (BEV) for autonomous driving, particularly where labeled data is limited. In the literature, Exponential Moving Average (EMA) has been used for adjustments of the weights of teacher network by the student network. However, the same induces catastrophic forgetting in the teacher network. In this work, we address this issue by introducing a novel concept of \textbf{Reflective Teacher} where the student is trained by both labeled and pseudo labeled data while its knowledge is progressively passed to the teacher through a regularizer to ensure retention of previous knowledge. Additionally, we propose Geometry Aware BEV Fusion (GA-BEVFusion) for efficient alignment of multi-modal BEV features, thus reducing the disparity between the modalities - camera and LiDAR. This helps to map the precise geometric information embedded among LiDAR points reliably with the spatial priors for extraction of semantic information from camera images. Our experiments on the nuScenes and Waymo datasets demonstrate: 1) improved performance over state-of-the-art methods in both fully supervised and semi-supervised settings; 2) \textit{Reflective Teacher} achieves equivalent performance with only 25\% and 22\% of labeled data for nuScenes and Waymo datasets respectively, in contrast to other fully supervised methods that utilize the full labeled dataset. 

\end{abstract}

%% file: include/introduction.tex
\section{Introduction}
\label{sec:Intro}

Visual perception is critical for applications like autonomous driving and robotics, as scene comprehension directly influences tasks such as path planning and control \cite{barrios2024deep, wu2024path}. Bird’s-eye-view (BEV) representation offers a clear depiction of scene objects, making it suitable for autonomous driving tasks utilizing sensors like LiDAR and cameras \cite{liu2023bevfusion, hu2023ea}. LiDAR offers precise yet sparse 3D point clouds, while cameras provide dense features but lack depth information. Multimodal fusion enhances accuracy and robustness by integrating the strengths of various sensors enhancing accuracy, robustness, in real-world applications \cite{das2024fisheye, dasgupta2022spatio, 10222711}. BEV integrates complimentary features from multiple sensors into a unified 3D space, but existing methods \cite{liu2023bevfusion, yin2024fusion} often result in feature misalignment \cite{song2024graphbev, song2023graphalign++} and sub-optimal performance. 
Typically, BEV architectures use a multi-view image encoder and a view transformation module \cite{yang2023bevformer} to convert perspective image features into BEV features. 
Moreover, autonomous driving methods rely on large labeled datasets \cite{caesar2020nuscenes, sun2020scalability}, which are costly to acquire. Semi-supervised learning \cite{liu2021unbiased} can be effectively leveraged here for training using minimal labeled data to generate high quality pseudo-labels as obtained by the teacher network.
In such training setup, active learning teacher-student schemes \cite{hekimoglu2024monocular, wang2023alwod} enhance supervision but risk catastrophic forgetting \cite{aljundi2018memory, chaudhry2018efficient}, as the teacher network updated by exponential moving average partially erases previously learned knowledge, leading to incorrect pseudo-label generation \cite{mi2022active}.

To resolve these issues, we summarize our contributions as follows:

\begin{itemize}





\item {Introducing \textbf{Reflective Teacher}, a BEV-based semi-supervised approach for multimodal 3D object detection, a first-of-its-kind approach to address catastrophic forgetting during the weight update process from the student network to the teacher. This approach also integrates an uncertainty measure into student network training, ensuring the teacher generates reliable pseudo-labels.}

\item {To seamlessly integrate multi sensor information using standard modality specific feature extractors such as BEVFormer V2 \cite{yang2023bevformer} and VoxelNet \cite{zhou2018voxelnet}, we introduce a novel fusion approach, GA-BEVFusion designed to address sensor specific feature alignment challenges thereby enhancing the reliability and robustness of the model.}




\item{Our method achieves state-of-the-art results on the nuScenes \cite{caesar2020nuscenes} and Waymo dataset \cite{sun2020scalability} in both fully supervised and semi-supervised setup that use only 25\% and 22\% labeled data respectively compared to fully supervised approaches.}

\end{itemize}







%% file: include/related_work.tex
\section{Related Work}

\subsection{Multimodal 3D object detection:}
As LiDAR features and camera features provide complementary information, current trends in multi-sensor fusion have indicated that merging these two modalities improves overall performance for 3D detection tasks. 
Some recent method \cite{hu2023ea} uses edge aware depth information to alleviate the “depth jump” problem
whereas some \cite{wang2023object} lift the 2D detector to 3D and utilize 2D objects as queries for 3D detection.
Different from these, CVFNet \cite{gu2022cvfnet} integrates both LiDAR and range view features in a progressive manner to reduce feature inconsistency. Additionally, feature-level fusion methods involve bounding box proposals of LiDAR features querying image features using a transformer \cite{bai2022transfusion,chen2023futr3d}, which are then concatenated back into the feature space. Recent advancements such as BEVFusion \cite{liu2023bevfusion,liang2022bevfusion} transform images and point clouds into a unified BEV space using attention mechanisms. The inclusion of temporal information in the framework, pioneered by LIFT \cite{zeng2022lift}, has been prompted by the capture of motion cues in adjacent frames \cite{cai2023bevfusion4d}. Without requiring inter-modality projection, \cite{cai2023objectfusion} incorporates multi-modal fusion at the object level across voxel, BEV, and image features, whereas \cite{yin2024fusion} at both local instance and global scene level.

\subsection{Semi-supervised object detection:}
Semi-supervised learning-based object detection methods harness the expanding pool of unlabeled data through approaches such as consistency regularization and pseudo-labeling. Consistency-based methods, such as SESS \cite{zhao2020sess}, employ diverse augmentations on unlabeled data to ensure consistent output between teacher and student predictions. Additionally, it incorporates the exponential moving average (EMA) proposed by Tarvainen et al. \cite{tarvainen2017mean} for updating the teacher model parameters. Pseudo-labeling based methods employ a teacher-student paradigm for initially training the teacher model on labeled data and subsequently ensure the student model's consistency with pseudo-labels, primarily emphasizing the pseudo-label quality. Wang et al. \cite{wang20213dioumatch} suggest dynamically filtering low-confidence pseudo-labels based on IoU learning, while Yin et al. \cite{yin2022semi} propose a clustering-based box voting module to handle multiple overlapping bounding boxes. To avoid low-quality prediction due to the inherent noise in pseudo-labels, Wang et al. \cite{wang2021combating} quantify region uncertainty and incorporate soft labels into training. Furthermore, Nozarian et al. \cite{nozarian2023reliable} use class-aware targets based on IoU value to reduce false negatives and query the teacher model to obtain the reliability score of the student-generated proposals. 

The present frameworks encounter various issues like the possibility of catastrophic forgetting during iterative pseudo-label generation, leading to loss of previous knowledge, and the generation of inaccurate pseudo-labels by the teacher model, which can negatively impact the training of the student network.  In contrast, our proposed architecture addresses catastrophic forgetting by using a regularizer to maintain crucial parameters and uphold previously acquired knowledge while also integrating uncertainty measures to enhance the reliability of pseudo-label generation through guidance from the teacher network.

%% file: include/architecture.tex
\section{Proposed Approach}

\begin{figure*}[!h]
    \centering
    \includegraphics[width=0.95\textwidth]{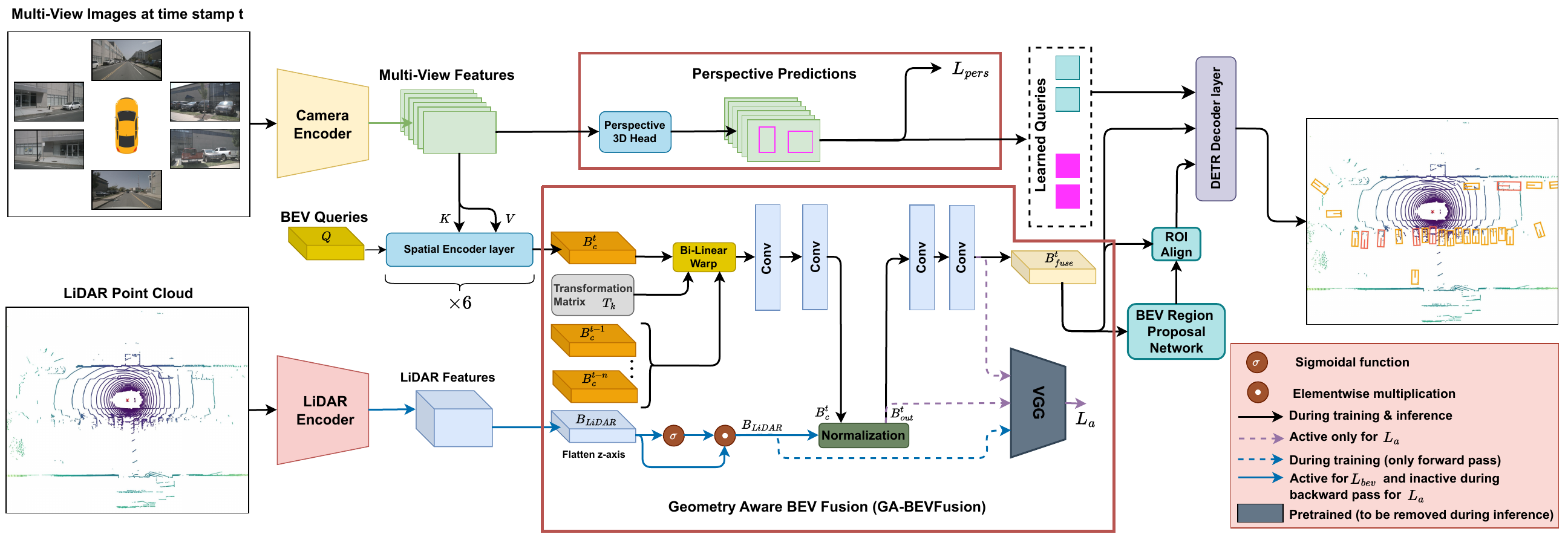}
    \captionsetup{font=small}
    \caption{
     The architecture of our proposed end-to-end multi-modal framework for 3D object detection in BEV. 
     }

    

    \label{fig:architecture}
    \vspace{-5mm}
\end{figure*}


 
\subsection{Overall architecture}
\label{subsec:Dual-Modality}

The architecture in Fig.\ref{fig:architecture} integrates multi-view camera images and LiDAR data in a unified BEV space. The camera encoder is based on InternImage  \cite{wang2023internimage}, and the LiDAR encoder on VoxelNet  \cite{zhou2018voxelnet}. Inspired by BEVFormer v2 \cite{yang2023bevformer}, we integrate perspective supervision to help the 2D backbone understand the 3D scene. The features from multi-view camera images are passed through the perspective 3D head to generate proposals in the perspective view by employing FCOS3D \cite{wang2021fcos3d} architecture. The spatial encoder uses cross-attention with deformable attention \cite{zhu2020deformable} to convert multi-camera images to the BEV plane. On the other hand, LiDAR features are projected into BEV space by flattening along the height dimension. Despite being in the same space, both LiDAR and camera BEV features can still exhibit spatial misalignment to some extent. To address this, the GA-BEVFusion module compensates for geometrical misalignment using an alignment loss $L_{a}$, resulting in robust BEV features.

The BEV head decoder, inspired by Deformable DETR, used in BEVFormer v2 \cite{yang2023bevformer}, includes a cross-attention layer with a deformable attention module using three inputs: the BEV feature map, query features, and reference points. Unlike BEVFormer v2, which uses projected bounding box centers as reference points, we use box centers from BEV RPN proposals for more accurate detection. The object queries in the BEV head generally rely on randomly initialized embeddings, which gradually learn to locate target objects over time. To expedite learning and improve accuracy, we filter region proposals from the perspective head through post-processing and use them as object queries for the decoder. The perspective loss \( L_{pers} \) complements the BEV loss \( L_{bev} \). Both detection heads are trained jointly with their respective losses and alignment loss as shown in Eq. (\ref{eq:total_loss}), balanced by coefficients \( \lambda \) and \( \gamma \).
%
%
%
%
\begin{equation}\label{eq:total_loss}
    L_{total} = L_{bev} + \lambda L_{pers} + \gamma L_{a}
\end{equation}
During training at each timestamp $t$, the model weights are updated through the backpropagation of the loss function $L_{total}$. The parameters of the LiDAR encoder are updated solely through the backpropagation of $L_{bev}$, while remaining inactive in case of $L_{a}$. 

\subsection{GA-BEVFusion}
\label{subsec:gabev}

Our model generates BEV features for the current timestamp $t$ using both spatial and temporal cues. A bi-linear warping strategy aligns previous BEV features $\{ B_c^{t-n}, \ldots, B_c^{t-1} \}$ with the current BEV feature $B_c^t$ using a transformation matrix $T_k^t = [R|t]$. The aligned features are concatenated with the current BEV feature along the channel dimension. Deformable convolutions then generate the enhanced BEV feature $B_c^t$ which improves understanding of scene dynamics. Additionally, the LiDAR BEV $B^{t}_{LiDAR}$ is updated using a sigmoidal function and element-wise multiplication with itself.


The GA-BEVFusion block employs deformable convolutional layers to enhance the resilience of BEV features against geometric transformations. This is achieved by dynamically selecting non-uniform locations within the receptive field by incorporating 2D offsets. The geometrical and positional information of the camera BEV features are aligned with the LiDAR BEV features by scaling normalized camera BEV features with variance $\sigma(B^{t}_{LiDAR})$ and shifting it by mean $\mu(B^{t}_{LiDAR})$. Consequently, each camera BEV feature, $B^{t}_{\text{c}}$, is modified to have the same mean and variance as that of LiDAR, $B^{t}_{LiDAR}$. The output feature of $B^{t}_{\text{c}}$ after normalization is given as follows:
\begin{equation}
B^{t}_{out} = 
\sigma(B^{t}_{LiDAR}) \left( \frac{B^{t}_{\text{c}} - \mu(B^{t}_{\text{c}})}{\sigma(B^{t}_{\text{c}})} \right) + \mu(B^{t}_{LiDAR})
\end{equation}
Finally, $B^{t}_{out}$ is passed through a series of deformable convolutions to produce fused BEV feature $B^{t}_{fuse}$. Supervision is done by calculating the alignment loss function ($L_{a}$) after passing $B_{out}^t$, $B_{fuse}^t$ and $B^{t}_{LiDAR}$ through a pre-trained VGG network, as shown in Eq. (\ref{eq:LA}).
\begin{equation}
\label{eq:LA}
\begin{split}
L_{a} = \sum_{i=1}^{k} \| \mu(f_i(B_{fuse}^t)) - \mu(f_i(B_{LiDAR}^t)) \|_2  \\
\quad + \sum_{i=1}^{k} \| \sigma(f_{i}(B_{fuse}^t)) - \sigma(f_{i}(B_{LiDAR}^t)) \|_2 + \\
 \| f(B_{out}^t) - f(B_{LiDAR}^t) \|_2
\end{split}
\end{equation} 
\noindent
where, each $f_i$ denotes an intermediate layer and $f$, the final layer in VGG-19 \cite{simonyan2014very}. In our implementation, we consider the features of layers relu1\_1, relu2\_1, relu3\_1 and relu4\_1 of the VGG-19 network to calculate the misalignment in both the distributions.


\subsection{Reflective Teacher Learning}
\label{subsec:ReflectiveTeacherLearning}

\begin{figure}[!t]
       \centering 
       \includegraphics[height=2.4in]{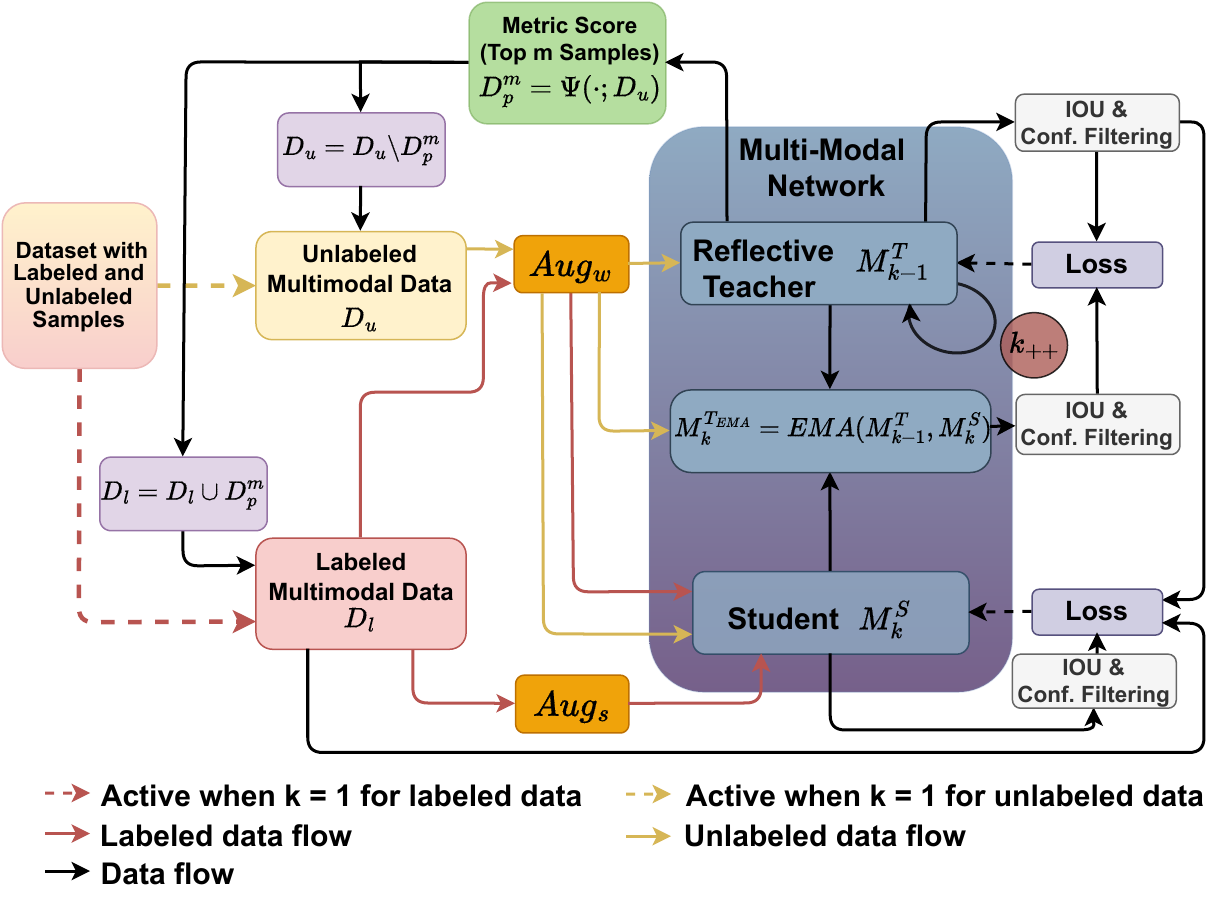}
        \captionsetup{font=small}
       \caption{Data flow sequence and interaction of teacher-student module in our proposed Reflective Teacher learning setup.}
       \vspace{12pt}
       \label{fig:fusion}
\end{figure}

The proposed architecture operates within a teacher-student framework, as shown in Fig. \ref{fig:fusion}, using multi-view camera images and LiDAR data. It aims to prevent catastrophic forgetting through reflective learning, ensuring previously gained information is preserved. In semi-supervised learning, the model leverages both labeled and unlabeled dataset, $D_l = \{x_l^i, y_l^i\}_{i=1}^{N_l}$, and  $D_u = \{x_u^i\}_{i=1}^{N_u}$ respectively. Here $y_l^i$ consists of category label $y_{cls}^i$ and bounding box coordinates $y_{reg}^i$. 
Robust initialization is essential for generating reliable pseudo-labels. Initially, the teacher is trained with a supervised dataset, as described in  Eq.~\eqref{eq:total_loss}, and the student parameters mirror the teacher's. The teacher model generates pseudo-labels from the unlabeled set. However, the accuracy of pseudo-label generation hinges on the diversity of the student model. Therefore, strongly ($Aug_s$) and weakly ($Aug_w$) augmented images are used as input for the student, while the teacher benefits from weakly augmented images during training.


Loss for the student network, $L_{stud}$, consists of supervised loss, $L_s$, and unsupervised loss, $L_u$. $L_s$ is defined as:
\vspace{-1mm}
\begin{equation}
\begin{split}
    L_s = \frac{1}{N_{l}} \sum_{i=1}^{N_{l}} ({L}_{cls}^{bev}(x_{l}^i, y_{cls}^i) + {L}_{reg}^{bev}(x_{l}^i, y_{reg}^i) + \\
    \lambda {L}_{pers}(x_{l}^i, y_{l}^i) + \gamma L_{a}(x_{l}^i))
\end{split}
\end{equation}
\noindent
where, ${L}_{cls}^{bev}$ and ${L}_{reg}^{bev}$ are the losses originating from RPN and ROI head, as detailed in Eq. (\ref{eq:bev_loss}). Here, $N_{l}$ is the total number of labeled samples, $y_{cls}$ and $y_{reg}$ are the class labels and bounding box coordinates respectively.
\begin{equation}
\begin{aligned}
\label{eq:bev_loss}
L_{cls}^{bev}(x_l^i, y_{cls}^i) = L_{cls}^{rpn}(x_l^i, y_{cls}^i) + L_{cls}^{roi}(x_l^i, y_{cls}^i) \\
L_{reg}^{bev}(x_l^i, y_{reg}^i) = L_{reg}^{rpn}(x_l^i, y_{reg}^i) + L_{reg}^{roi}(x_l^i, y_{reg}^i)
\end{aligned}
\end{equation}
The unsupervised loss, $L_u$, is given by (\ref{eq:unsup}), where $N_{\text{u}}$ is the total number of unlabeled samples, $\hat{y}_{cls}$ and $\hat{y}_{reg}$ are the respective class labels
and bounding box coordinates of the pseudo-label $\hat{y}_u$ generated by the teacher model, while $L_{cls}^{bev}$ and $L_{reg}^{bev}$ are the same as in Eq.~\eqref{eq:bev_loss}.
\vspace{-1mm}
\begin{equation}
\begin{split}
\label{eq:unsup}
    L_u = \frac{1}{N_{u}} \sum_{i=1}^{N_{u}} (L_{cls}^{bev}(x_{u}^i, \hat{y}_{cls}^i) + L_{reg}^{bev}(x_{u}^i, \hat{y}_{reg}^i) + \\
    \lambda{L}_{pers}(x_{u}^i, \hat{y}_{u}^i) + \gamma L_{a}(x_{u}^i))
\end{split}
\end{equation} 
The final student network loss function is defined by Eq. (\ref{eq:stloss}). Here, $\kappa$ balances the trade off between supervised and unsupervised loss.
\begin{equation}\label{eq:stloss}
    L_{stud} = L_s + \kappa L_u
\end{equation} 
At any $k^{th}$ iteration during the training, the labeled set is gradually augmented with pseudo-labels derived from the teacher model $M_{k-1}^T$. Mi et al. \cite{mi2022active} demonstrated an effective strategy to augment the labeled set by measuring three active sampling metrics on unlabeled data: difficulty, information, and diversity, which are then normalized and aggregated to a single sampling score $\beta$. The function $\Psi$ ranks the unlabeled set $D_u^{k-1}$ based on the sampling score $\beta $ and returns the top-m data points with pseudo-labels, $D_p^m$. This is expressed as $D_p^m = \Psi(:,D_u^{k-1})$, while $D_l^k=D_l^{k-1} \cup D_p^m$ and $D_u^k = D_u^{k-1} $\textbackslash$ D_p^m$ are the updated labeled and unlabeled set respectively.

The student model at the present iteration $k$, $M_k^S$ is then trained on this modified labeled and unlabeled data by Eq.~\eqref{eq:stloss}. The parameters of the teacher model $M_k^{T_{EMA}}$ is updated with the help of student parameters through \textit{Exponential Moving Average} (EMA),  $\theta_k^{T_{EMA}} \leftarrow \alpha\theta_{k-1}^T + (1 - \alpha)\theta_k^S$, $\alpha$ being the EMA coefficient, $\theta_{k-1}^T$ and $\theta_k^S$, the weights of the previous teacher model $M_{k-1}^T$ and student model $M_k^S$ respectively. However, EMA update can affect the previously learned weights, causing catastrophic forgetting. To address this, a reflective learning ability is essential for the model to incrementally acquire and integrate new knowledge, thus preventing alterations to critical parameters and preserving predictions at each observed data point. Similar to \cite{aljundi2018memory}, we calculate the gradient of the previously learned teacher network $M_{k-1}^T$ with respect to the parameter $\theta_{ij}$ (weight between the neurons $n_i$ and $n_j$) at each sample point $x_u$ of the previous unlabeled set to measure the sensitivity of learned function to parameter changes. The gradients are accumulated over the unlabeled data points to obtain importance weight $\Phi_{ij}$ for parameter $\theta_{ij}$ as:
\begin{equation}
\Phi_{ij}^k = \mathbb{E}_{x_u \in D_u^{k-1}} [ \frac{\partial \|  M_{k-1}^T(x_u; \theta) \|_2^2}{\partial \theta_{ij}}]
\end{equation}
To address potential discrepancies between pseudo-labels generated by the $M_k^{T_{EMA}}$ and those produced by the previous teacher network $M_{k-1}^T$ when unlabeled data is processed, a regularizer is incorporated into the loss function, as depicted in Eq. (\ref{eq:tcloss}), to penalize the changes in the parameters.
\begin{equation}
\begin{aligned}\label{eq:tcloss}
L_{ref}^k(\theta) = L(M_k^{T_{EMA}}(x_u, \theta_k^{T_{EMA}}),M_{k-1}^T(x_u, \theta^T_{k-1})) \\ + 
\eta \sum_{i,j} \Phi_{ij}^k (\theta_{ij} - \theta_{ij}^{k-1})^2
\end{aligned}
\end{equation} 
Here, $\eta$ is a hyperparameter for the regularizer, $\theta_{ij}$ and $\theta_{ij}^{k-1}$ are the parameter of $M_k^{T_{EMA}}$ and $M_{k-1}^T$ respectively. $L(:,:)$ calculates the L1 loss \cite{deng2021voxel} between $M_k^{T_{EMA}}(x_u, \theta_k^{T_{EMA}})$ and $M_{k-1}^T(x_u, \theta^T_{k-1})$ for all $x_u\in D_u^{k}$. The weights of the teacher network, $M_k^T$, are updated iteratively through backpropagation of the loss function $L_{ref}^k(\theta)$ until reaching the maximum iteration, resulting in a well-trained teacher network.

\subsection{Uncertainty Measure of Regions}
\label{subsec:UncertaintyMoR}

When dealing with unlabeled data, inaccurate pseudo-labels from the teacher model may introduce noise, impeding the effectiveness of the student model. Therefore, it becomes imperative to devise a way to mitigate the noise present in the pseudo-labels. For an unlabeled image, the BEV RPN (Region Proposal Network) \cite{deng2021voxel} of the student model generates a set of region proposals from which a certain number of region proposals {${\hat{r}}_i$} are randomly sampled for subsequent classification and regression. Each region proposal {${\hat{r}}_i$} is labeled based on its maximum overlap with the pseudo-labels generated by the teacher network. The IoU $I_i$, as expressed in Eq.(\ref{eq:iou}), measures the maximum spatial overlap between a proposal and the pseudo-labels. Higher $I_i$ value signifies increased certainty in the proposal's classification and localization accuracy. 

\begin{equation}\label{eq:iou}
    I_i = IoU_{max}({\hat{r}}_i, \hat{y}_u)
\end{equation}
\noindent
where $\hat{y}_u$ belongs to the pseudo-label set of the teacher network. If $I_i$ exceeds a predefined threshold $\Delta$ it is considered as a positive proposal, and the corresponding label \{{$b_i$},{$c_i$},{$s_i$}\} from the pseudo-label set is assigned. Otherwise, it is labeled as a negative proposal, with category {$c_i$} as background, and no assignment for bounding box {$b_i$} or probability score {$s_i$}. But for the unlabeled data if the assigned category {$c_i$} for the proposal ${\hat{r}}_i$ ends up being incorrect, the region proposal ${\hat{r}}_i$ may end up uncertain. This uncertainty makes region proposals ambiguous. To address this issue, we introduce a metric to measure the uncertainty associated with region proposals.

Incorrect pseudo-label categories can result in mislabeled proposals ${\hat{r}}_i$. The classification probability score {$s_i$} serves as a reliable indicator of self-confidence, making it suitable for measuring the inherent noise. Based on this, we use both {$s_i$} and {$I_i$} to evaluate uncertainty for the proposal, as shown below :
\begin{equation}
u_i = 
    \begin{cases}
    1 - (s_i \cdot I_i)^{\beta'}, & \text{if } I_i > \Delta \\
    0, & \text{otherwise}
\end{cases}
\quad \beta' = \frac{1}{1 + e^{-{\beta}}}
\end{equation}
When both $s_i$ and $I_i$ are close to 1, it indicates the region proposal ${\hat{r}}_i$ is relatively certain about its assigned label $c_i$ and $u_i$ is thus close to 0. The negative proposals are presumed to be less prone to noise as they are considered certain to the model. So we set their uncertainty to 0. By multiplying (1-$u_i$) with the unsupervised RPN loss, we regularize the training process to encourage the model to be less confident in uncertain regions, thereby preventing overfitting to noisy or uncertain data. The new unsupervised RPN loss in Eq.~\eqref{eq:unsup} is modified as follows:
\begin{equation}
    L_{reg}^{rpn}(x_u^i, \hat{y}_{reg}^i) = \sum_{j=1}^{k_{\text{i}}} (1-u_{\hat{r}_j})L_{reg}^{rpn}({\hat{r}_j}, \hat{y}_{\hat{r}_j})
\end{equation} 
where, $\hat{y}_{\hat{r}_j}$ is the assigned label of the region proposal ${\hat{r}_j}$.

%% file: include/results.tex
\section{Experimentation Details}

\begin{table*}[!ht]
    \centering
    \scalebox{.67}{
     \begin{tabular}{l|l|l|l|l|l|l|l|l|l|l|l|l|l}
    \hline
        Methods & Modality & mAP & NDS & Car & Truck & CV & Bus & Trailer & Barrier & Motor & Bike & Ped & TC  \\ \hline
        CenterPoint-Pillar \cite{yin2021center} & L   & 50.3 & 60.2 & - & - & - & - & - & - & - & - & - & -  \\ 
        CenterPoint-Voxel \cite{yin2021center} & L  & 56.4 & 64.8 & - & - & - & - & - & - & - & - & - & -  \\ 
        TransFusion-L \cite{bai2022transfusion} & L  & 65.1 & 70.1 & 86.5 & 59.6 & 25.4 & 74.4 & 42.2 & 74.1 & 72.1 & 56.0 & 86.6 & 74.1  \\ \hline
        BEV-IO-R50 \cite{zhang2023bev}& C  & 36.8 & 49.3 & - & - & - & - & - & - & - & - & - & -  \\ 
        DETR3D \cite{wang2022detr3d} & C  & 34.6 & 42.5 & - & - & - & - & - & - & - & - & - & -  \\ 
        BEVFormer \cite{li2022bevformer} & C &  41.6 & 51.7 & - & - & - & - & - & - & - & - & - & -  \\ 
        Fast-BEV(R101) \cite{huang2023fast} & C &  41.3 & 53.5 & - & - & - & - & - & - & - & - & - & -  \\ \hline
        TransFusion  \cite{bai2022transfusion} & LC & 67.3 & 71.2 & 87.6 & 62.0 & 27.4 & 75.7 & 42.8 & 73.9 & 75.4 & 63.1 & 87.8 & 77.0  \\ 
        FUTR3D \cite{chen2023futr3d} & LC  & 64.2 & 68.0 & 86.3 & 61.5 & 26.0 & 71.9 & 42.1 & 64.4 & 73.6 & 63.3 & 82.6 & 70.1  \\ 
        BEVFusion \cite{liu2023bevfusion} & LC & 68.5 & 71.4 & 89.2 & 64.6 & 30.4 & 75.4 & 42.5 & 72.0 & 78.5 & 65.3 & 88.2 & 79.5  \\ 
        BEVFusion \cite{liang2022bevfusion} & LC  & 69.6 & 72.1 & 89.1 & 66.7 & 30.9 & 77.7 & 42.6 & 73.5 & 79 & 67.5 & 89.4 & 79.3  \\ 
        DeepInteraction \cite{yang2022deepinteraction} & LC  & 69.9 & 72.6 & 88.5 & 64.4 & 30.1 & 79.2 & 44.6 & \textit{76.4} & 79 & 67.8 & 88.9 & 80  \\ 
        BEVFusion4D-S  \cite{cai2023bevfusion4d} & LC  & 70.9 & 72.9 & \textit{89.8} & \textit{69.5} & \textit{32.6} & \textit{80.6} & \textit{46.3} & 71.0 & \textit{79.6} & \textit{70.3} & \textit{89.5} & 80.3  \\ 
        ObjectFusion  \cite{cai2023objectfusion} & LC  & 69.8 & 72.3 & 89.7 & 65.6 & 32.0 & 77.7 & 42.8 & 75.2 & 79.4 & 65.0 & 89.3 & \textit{81.1}  \\ 
        EA-LSS \cite{hu2023ea} & LC & \textit{71.2} & \textit{73.1} & - & - & - & - & - & - & - & - & - & - \\
        \textbf{\textcolor{blue}{Ours}} & \textbf{\textcolor{blue}{LC}} & \textbf{\textcolor{blue}{72.6}} & \textbf{\textcolor{blue}{74.1}} & \textbf{\textcolor{blue}{90.24}} & \textbf{\textcolor{blue}{70.3}} & \textbf{\textcolor{blue}{33.87}} & \textbf{\textcolor{blue}{80.1}} & \textbf{\textcolor{blue}{50.1}} & \textbf{\textcolor{blue}{77.4}} & \textbf{\textcolor{blue}{80.7}} & \textbf{\textcolor{blue}{67.7}} & \textbf{\textcolor{blue}{93.2}} & \textbf{\textcolor{blue}{82.8}}  \\ \hline
        \textbf{\textcolor{blue}{Improvement}} & \textbf{\textcolor{blue}{vs. LC}} & \textbf{\textcolor{blue}{+0.4\%}}  & \textbf{\textcolor{blue}{+1.0\%}} & \textbf{\textcolor{blue}{+0.44\%}}  & \textbf{\textcolor{blue}{+0.8\%}} & \textbf{\textcolor{blue}{+1.27\%}}  & \textbf{\textcolor{blue}{-0.5\%}}  & \textbf{\textcolor{blue}{+3.8\%}} & \textbf{\textcolor{blue}{+1.0\%}}  & \textbf{\textcolor{blue}{+1.1\%}} & \textbf{\textcolor{blue}{-2.6\%}} & \textbf{\textcolor{blue}{+3.7\%}}  & \textbf{\textcolor{blue}{+1.7\%}} \\
        \hline

        BEVFusion4D  \cite{cai2023bevfusion4d} & LCT & \textit{72.0} & \textit{73.5} & \textit{90.6} & \textit{70.3} & \textit{32.9} & \textit{81.5} & \textit{47.1} & \textit{71.6} & \textit{81.5} & \textit{73.0} & \textit{90.2} & \textit{80.9}  \\ 
        \textbf{\textcolor{blue}{Ours}} & \textbf{\textcolor{blue}{LCT}} & \textbf{\textcolor{blue}{74.6}} & \textbf{\textcolor{blue}{75.4}} & \textbf{\textcolor{blue}{92.2}} & \textbf{\textcolor{blue}{70.7}} & \textbf{\textcolor{blue}{34.7}} & \textbf{\textcolor{blue}{82.1}} & \textbf{\textcolor{blue}{58.1}} & \textbf{\textcolor{blue}{77.9}} & \textbf{\textcolor{blue}{82.1}} & \textbf{\textcolor{blue}{73.3}} & \textbf{\textcolor{blue}{92.5}} & \textbf{\textcolor{blue}{82.4}}\\
        \hline
        \textbf{\textcolor{blue}{Improvement}} & \textbf{\textcolor{blue}{vs. LCT}} & 
        \textbf{\textcolor{blue}{+2.6\%}}  & \textbf{\textcolor{blue}{+1.9\%}} & \textbf{\textcolor{blue}{+1.6\%}}  & \textbf{\textcolor{blue}{+0.4\%}} & \textbf{\textcolor{blue}{+1.8\%}}  & \textbf{\textcolor{blue}{+0.6\%}}  & \textbf{\textcolor{blue}{+11.0\%}} & \textbf{\textcolor{blue}{+6.3\%}}  & \textbf{\textcolor{blue}{+0.6\%}} & \textbf{\textcolor{blue}{+0.3\%}} & \textbf{\textcolor{blue}{+2.3\%}}  & \textbf{\textcolor{blue}{+1.5\%}} \\
        \hline
    \end{tabular}
    
    }
     \label{tab:eval_val}
\end{table*}

\begin{table*}[!ht]
    \centering
    \scalebox{.67}{
     \begin{tabular}{l|l|l|l|l|l|l|l|l|l|l|l|l|l}
    \hline
     Methods & Modality & mAP & NDS & Car & Truck & CV & Bus & Trailer & Barrier & Motor & Bike & Ped & TC  \\ \hline
       
        TransFusion-L \cite{bai2022transfusion} & L & 65.5 & 70.2 & 86.2 & 56.7 & 28.2 & 66.3 & 58.8 & 78.2 & 68.3 & 44.2 & 86.1 & 82.0  \\ 
        PointPillar \cite{lang2019pointpillars} & L & 40.1 & 55.0 & 76.0 & 31.0 & 11.3 & 32.1 & 36.6 & 56.4 & 34.2 & 14.0 & 64.0 & 45.6  \\ 
        CenterPoint \cite{yin2021center} & L & 60.3 & 67.3 & 85.2 & 53.5 & 20.0 & 63.6 & 56.0 & 71.1 & 59.5 & 30.7 & 84.6 & 78.4  \\ \hline
        DETR3D \cite{wang2022detr3d} & C & 41.2 & 47.9 & - & - & - & - & - & - & - & - & - & -  \\ 
        BEVFormer \cite{li2022bevformer} & C  & 48.1 & 56.9 & - & - & - & - & - & - & - & - & - & -  \\ 
        BEVFormer v2 \cite{yang2023bevformer} & C  & 55.6 & 63.4 & - & - & - & - & - & - & - & - & - & -  \\ \hline
        BEVFusion \cite{{liu2023bevfusion}} & LC & 70.2 & 72.9 & 88.6 & 60.1 & 39.3 & 69.8 & 63.8 & 80.0 & 74.1 & 51.0 & 89.2 & 86.5  \\ 

        BEVFusion \cite{liang2022bevfusion} & LC & 71.3 & 73.3 & 88.1 & 60.9 & 34.4 & 69.3 & 62.1 & 78.2 & 72.2 & 52.2 & 89.2 & 86.7  \\ 
        DeepInteraction \cite{yang2022deepinteraction} & LC & 70.8 & 73.4 & 87.9 & 60.2 & 37.5 & 70.8 & 63.8 & {80.4} & 75.4 & 54.5 & 91.7 & 87.2  \\
        BEVFusion4D-S \cite{cai2023bevfusion4d} & LC & 71.9 & 73.7 & 88.8 & 64.0 & 38.0 & 72.8 & 65.0 & 79.8 & 77.0 & 56.4 & 90.4 & 87.1  \\ 
        TransFusion \cite{bai2022transfusion} & LC & 68.9 & 71.7 & 87.1 & 60.0 & 33.1 & 68.3 & 60.8 & 78.1 & 73.6 & 52.9 & 88.4 & 86.7  \\ 
        ObjectFusion \cite{cai2023objectfusion} & LC & 71.0 & 73.3 & {89.4} & 59.0 & {40.5} & 71.8 & 63.1 & 76.6 & 78.1 & 53.2 & 90.7 & 87.7  \\ 
         MV2DFusion-e* \cite{wang2023object} & LC & {\em 77.9} &\em 78.8 & \em 91.1 & \em 69.7 & \em 45.7 & 76.7 & \em 70.4 & 83.0 & \em 87.1 & \em 72.2 & 93.3 & 90.3 \\ 
        EA-LSS* \cite{hu2023ea} & LC & 76.6 & 77.6 & 90.2 & 67.1 & 43.9 & 76.7 & 69.1 & \em 84.1 & 85.9 & 66.6 & 91.3 & \em 91.2 \\
        IS-Fusion \cite{yin2024fusion} & LC & 76.5 & 77.4 & 89.8 & 67.8 & 44.5 & \em 77.6 & 68.3 & 81.8 & 85.3 & 65.6 & \em 93.4 & 91.1 \\ 
        \textbf{\textcolor{blue}{Ours}} & \textbf{\textcolor{blue}{LC}} & \textbf{\textcolor{blue}{76.8}} & \textbf{\textcolor{blue}{78.3}} & \textbf{\textcolor{blue}{91.2}} & \textbf{\textcolor{blue}{70.3}} & \textbf{\textcolor{blue}{37.4}} & \textbf{\textcolor{blue}{82.5}} & \textbf{\textcolor{blue}{69.5}} & \textbf{\textcolor{blue}{77.6}} & \textbf{\textcolor{blue}{88.2}} & \textbf{\textcolor{blue}{73.1}} & \textbf{\textcolor{blue}{95.3}} & \textbf{\textcolor{blue}{83.1}}  \\ \hline
        \textbf{\textcolor{blue}{Improvement}} & \textbf{\textcolor{blue}{vs. LC}} & \textbf{\textcolor{blue}{-1.1\%}}  & \textbf{\textcolor{blue}{-0.5\%}} & \textbf{\textcolor{blue}{+0.1\%}}  & \textbf{\textcolor{blue}{+0.6\%}} & \textbf{\textcolor{blue}{+8.3\%}}  & \textbf{\textcolor{blue}{+4.9\%}}  & \textbf{\textcolor{blue}{-0.9\%}} & \textbf{\textcolor{blue}{-6.5\%}}  & \textbf{\textcolor{blue}{+1.1\%}} & \textbf{\textcolor{blue}{+0.9\%}} & \textbf{\textcolor{blue}{+1.9\%}}  & \textbf{\textcolor{blue}{-8.1\%}} 
        \\
        \hline
        LIFT \cite{zeng2022lift} & LCT & 65.1 & 70.2 & 87.7 & 55.1 & 29.4 & 62.4 & 59.3 & 69.3 & 70.8 & 47.7 & 86.1 & 83.2  \\ 
        BEVFusion4D \cite{cai2023bevfusion4d} & LCT & \textit{73.3} & \textit{74.7} & \textit{89.7} & \textit{65.6} & \textit{41.1} & \textit{72.9} & \textit{66.0} & \textit{81.0} & \textit{79.5} & \textit{58.6} & \textit{90.9} & \textit{87.7}  \\ 
        \textbf{\textcolor{blue}{Ours}} &
        \textbf{\textcolor{blue}{LCT}} &  \textbf{\textcolor{blue}{79.3}} & \textbf{\textcolor{blue}{80.4}} & \textbf{\textcolor{blue}{92.7}} & \textbf{\textcolor{blue}{71.5}} & \textbf{\textcolor{blue}{47.1}} & \textbf{\textcolor{blue}{77.9}} & \textbf{\textcolor{blue}{70.4}} & \textbf{\textcolor{blue}{84.6}} & \textbf{\textcolor{blue}{88.5}} & \textbf{\textcolor{blue}{73.6}} & \textbf{\textcolor{blue}{95.2}} & \textbf{\textcolor{blue}{91.6}} \\
        \hline
        \textbf{\textcolor{blue}{Improvement}} & \textbf{\textcolor{blue}{vs. LCT}} & \textbf{\textcolor{blue}{+6.0\%}}  & \textbf{\textcolor{blue}{+5.7\%}} & \textbf{\textcolor{blue}{+3.0\%}}  & \textbf{\textcolor{blue}{+5.9\%}} & \textbf{\textcolor{blue}{+6.0\%}}  & \textbf{\textcolor{blue}{+5.0\%}}  & \textbf{\textcolor{blue}{+4.4\%}} & \textbf{\textcolor{blue}{+3.6\%}}  & \textbf{\textcolor{blue}{+9.0\%}} & \textbf{\textcolor{blue}{+15.0\%}} & \textbf{\textcolor{blue}{+4.3\%}}  & \textbf{\textcolor{blue}{+3.9\%}} \\                                    
        \hline
    \end{tabular}
    }
  \captionsetup{font=small}
 \caption{Quantitative comparison of two GA-BEV Fusion variants (LC and LCT) with SOTA 3D object detection methods in multimodal settings on the nuScenes dataset \cite{caesar2020nuscenes} val (top) and test (bottom) sets. ``L'',``C'' and ``T'' denote LiDAR, Camera and Temporal respectively. Overall improvements for LCT and LC are highlighted in blue and the second highest mAPs and NDSs are marked in \textit{italics}. `*' indicates values from the nuScenes leaderboard.}
     \label{tab:nuscenes}
\end{table*}

\begin{table*}[h!]
\centering
 \scalebox{.67}{
\begin{tabular}{p{0.85in}|l|cc|cc|cc|cc}
\hline 
& Methods & \multicolumn{2}{c|}{mAP/mAPH} & \multicolumn{2}{c|}{Vehicle AP/APH} & \multicolumn{2}{c|}{Pedestrian AP/APH} & \multicolumn{2}{c}{Cyclist AP/APH} \\ 
 & &  L1       & L2       & L1         & L2        & L1          & L2          & L1        & L2 \\ \hline
\multirow{8}{*}{Waymo Val set}& CenterPoint-Voxel \cite{yin2021center} & 74.4/71.7 & 68.2/65.8 & 79.0/78.5 & 79.0/78.5 & 76.6/70.5 & 72.3/71.1 & 68.2/65.8 & 66.2/65.7 \\ 
 & Pillarnet-34  \cite{shi2022pillarnet}    & 77.3/74.6 & 70.9/68.4 & 79.0/78.5 & 79.0/78.5 & 80.5/70.4 & 72.2/71.2 & 72.3/71.1 & 72.2/66.1 \\ 
& FSD      \cite{fan2023super}         & 79.4/77.1 & 72.7/70.5 & 79.5/79.0 & 79.5/\textit{79.0} & 80.5/70.4 & 74.4/69.4 & 75.3/74.1 & 73.7/72.1 \\ 
& Centerformer   \cite{zhou2022centerformer}   & 75.3/72.9 & 71.7/68.9 & 75.0/74.4 & 69.9/69.4 & 78.6/73.0 & 74.7/73.7 & 76.8/73.0 & 69.9/69.4 \\ 
& Clusterformer  \cite{pei2023clusterformer}   & 81.4/79.0 & 74.6/72.3 & 79.8/77.9 & 75.7/70.7 & 84.4/79.0 & 80.0/77.6 & 80.0/78.7 & 77.4/76.6 \\ 
& Clusterformer-3f \cite{pei2023clusterformer}  & \em 83.3/81.7 & \em 77.7/76.2 & \em 81.4/79.3 & \textit{79.7}/76.1 & \em 85.9/82.7 & \em 82.5/81.6 & \em 82.5/81.6 & \em 79.6/78.7 \\ 
\hhline{~---------}
& \textcolor{blue}{\textbf{Ours}} & \textcolor{blue}{\textbf{85.5/83.2}} & \textcolor{blue}{\textbf{78.6/77.5}} & \textcolor{blue}{\textbf{82.8/81.2}} & \textcolor{blue}{\textbf{75.2/74.5}} & \textcolor{blue}{\textbf{86.7/83.4}} & \textcolor{blue}{\textbf{79.8/78.6}} & \textcolor{blue}{\textbf{86.8/85.3}} & \textcolor{blue}{\textbf{80.8/79.3}}\\
\hhline{~---------}
& \textcolor{blue}{\textbf{Improvement}} & \textcolor{blue}{\textbf{+2.2/+1.5}} & \textcolor{blue}{\textbf{+0.9/+1.3}} & \textcolor{blue}{\textbf{+1.4/+1.9}} & \textcolor{blue}{\textbf{-4.5/-4.5}} & \textcolor{blue}{\textbf{+0.8/+0.7}} & \textcolor{blue}{\textbf{-2.7/-3.0}} & \textcolor{blue}{\textbf{+4.3/+3.7}} & \textcolor{blue}{\textbf{+1.2/+0.6}}\\

\hline
\multirow{6}{*}{Waymo Test set} & PV-RCNN++   \cite{shi2023pv}     & 77.9/75.6 & 72.4/70.1 & 81.6/\em 81.2 & \em 81.6/81.2 & 78.1/70.2 & 74.1/69.0 & 71.9/70.7 & 69.2/68.1 \\ 
& Pillarnet-34  \cite{shi2022pillarnet}   & 78.1/75.9 & 71.7/69.7 & 79.3/78.8 & 79.3/78.8 & 81.3/76.3 & \em 77.3/72.7 & 73.7/72.7 & 71.2/70.2 \\ 
& AFDetV2-lite \cite{hu2022afdetv2}    & 77.5/75.2 & 72.2/70.0 & 80.5/80.0 & 80.5/80.0 & 80.5/78.0 & 73.0/72.6 & 72.4/71.2 & 68.7/67.4 \\ 
& Clusterformer  \cite{pei2023clusterformer}  & \em 81.1/78.9 & \em 75.0/73.0 & \textit{83.4}/78.3 & 73.1/72.8 & \em 83.4/78.3 & 76.6/71.9 & \em 78.4/77.1 & \em 75.6/77.1 \\ 
\hhline{~---------}
& \textcolor{blue}{\textbf{Ours}} & \textcolor{blue}{\textbf{85.9/84.1}} & \textcolor{blue}{\textbf{78.4/75.9}} & \textcolor{blue}{\textbf{84.8/83.9}} & \textcolor{blue}{\textbf{78.4/77.5}} & \textcolor{blue}{\textbf{87.9/83.9}} & \textcolor{blue}{\textbf{79.9/77.0}} & \textcolor{blue}{\textbf{85.1/84.6}} & \textcolor{blue}{\textbf{76.8/78.1}} \\
\hhline{~---------}
& \textcolor{blue}{\textbf{Improvement}} & \textcolor{blue}{\textbf{+4.8/+5.2}} & \textcolor{blue}{\textbf{+3.4/+2.9}} & \textcolor{blue}{\textbf{+1.4/+2.7}} & \textcolor{blue}{\textbf{-3.2/-3.7}} & \textcolor{blue}{\textbf{+4.5/+5.6}} & \textcolor{blue}{\textbf{+2.6/+4.3}} & \textcolor{blue}{\textbf{+6.7/+7.5}} & \textcolor{blue}{\textbf{+1.2/+1.0}} \\
\hline
\end{tabular}%
}
\vspace{-2mm}
\captionsetup{font=small}
\caption{Quantitative comparisons of different SOTA models on Waymo Val and Test set \cite{sun2020scalability}. The second highest mAPs and mAPH are marked in \textit{italics}.}
\label{tab:waymo_val}
\vspace{-3mm}
\end{table*}

\begin{table*}[!ht]
    \centering
    \scalebox{.72}{
    \begin{tabular}{l|cc|cc|cc|cc|cc|cc|cc|cc}
    \hline
     \multirow{2}{*}{Methods} & \multicolumn{8}{c|}{nuScenes} & \multicolumn{8}{c}{Waymo} \\
    \hhline{~----------------}
       & 5\% S & $\Delta\%$  & $10\%$ S & $\Delta\%$ & $20\%$ S & $\Delta\%$ & $25\%$ S & $\Delta\%$ &  5\% S & $\Delta\%$  & $10\%$ S & $\Delta\%$ & $20\%$ S & $\Delta\%$ & $22\%$ S & $\Delta\%$   \\ \hline
        Supervised  & 23.12 & - & 25.25 & - & 32.58 & - & 35.66 & - & 
        25.5 & - & 27.06 & - & 34.91 & - & 38.73 & -   \\ 
        Unbiased-Teacher  \cite{liu2021unbiased} & 39.34 & 7.38 & 42.21 & 8.43 & 52.75 & 7.48 & 55.94 & 8.19  & 
        43.94 & 6.81 & 45.65 & 7.67 & 56.56 & 7.59 & 59.38 & 10.01 \\ 
        Active Teacher  \cite{mi2022active} & 42.54 & 6.76 & 49.53 & 7.1 & 63.21 & 7.69 & 67.37 & 8.26  
        & 45.82 & 7.61 & 53.49 & 7.41 & 69.39 & 9.09 & 71.5 & 8.12  \\ \hline
        \textbf{\textcolor{blue}{Reflective Teacher}}  & \textbf{\textcolor{blue}{47.23}} & \textbf{\textcolor{blue}{0.4}} & \textbf{\textcolor{blue}{56.16}} & \textbf{\textcolor{blue}{0.53}} & \textbf{\textcolor{blue}{72.86}} & \textbf{\textcolor{blue}{0.79}} & \textbf{\textcolor{blue}{79.46}} & \textbf{\textcolor{blue}{0.86}} & 
        \textbf{\textcolor{blue}{55.4}} & \textbf{\textcolor{blue}{0.71}} & \textbf{\textcolor{blue}{62.31}} & \textbf{\textcolor{blue}{0.79}} & \textbf{\textcolor{blue}{80.23}} & \textbf{\textcolor{blue}{0.85}} & \textbf{\textcolor{blue}{85.31}} & \textbf{\textcolor{blue}{0.91}} \\ \hline
    \end{tabular}
    }
    \vspace{-1mm}
    \captionsetup{font=small}
     \caption{ Evaluation results on mAP value on the test set of nuScenes  \cite{caesar2020nuscenes} and Waymo \cite{sun2020scalability} dataset (L1) with only certain data percentages using the proposed semi-supervised setup and multimodal architecture. It also includes catastrophic forgetting percentages for the respective supervision denoted by ``S".}
     \label{tab:eval_data_percentage_nuscenes}
\end{table*}

\subsection{Evaluation metrics and results}
\noindent
\textbf{On nuScenes Dataset: }
We evaluate our model on the nuScenes dataset \cite{caesar2020nuscenes} using mean Average Precision (mAP) and nuScenes detection scores (NDS). Table \ref{tab:nuscenes} shows our model achieves 74.6\% mAP and 75.4\% NDS on the validation set, and 79.3\% mAP and 80.4\% NDS on the test set surpassing the previous SOTA models.

In Table \ref{tab:eval_data_percentage_nuscenes}, we see an improvement in mAP values compared to the supervised model. Also we show that our model, trained on only 25\% labeled data, achieves equivalent result to the fully labeled dataset (refer to Table \ref{tab:nuscenes}), demonstrating the effectiveness of our semi-supervised approach. The proposed Reflective-Teacher model outperforms other semi-supervised methods across various percentages of labeled data.

\begin{figure*}[!h]
    \centering
    \includegraphics[width=0.75\textwidth]{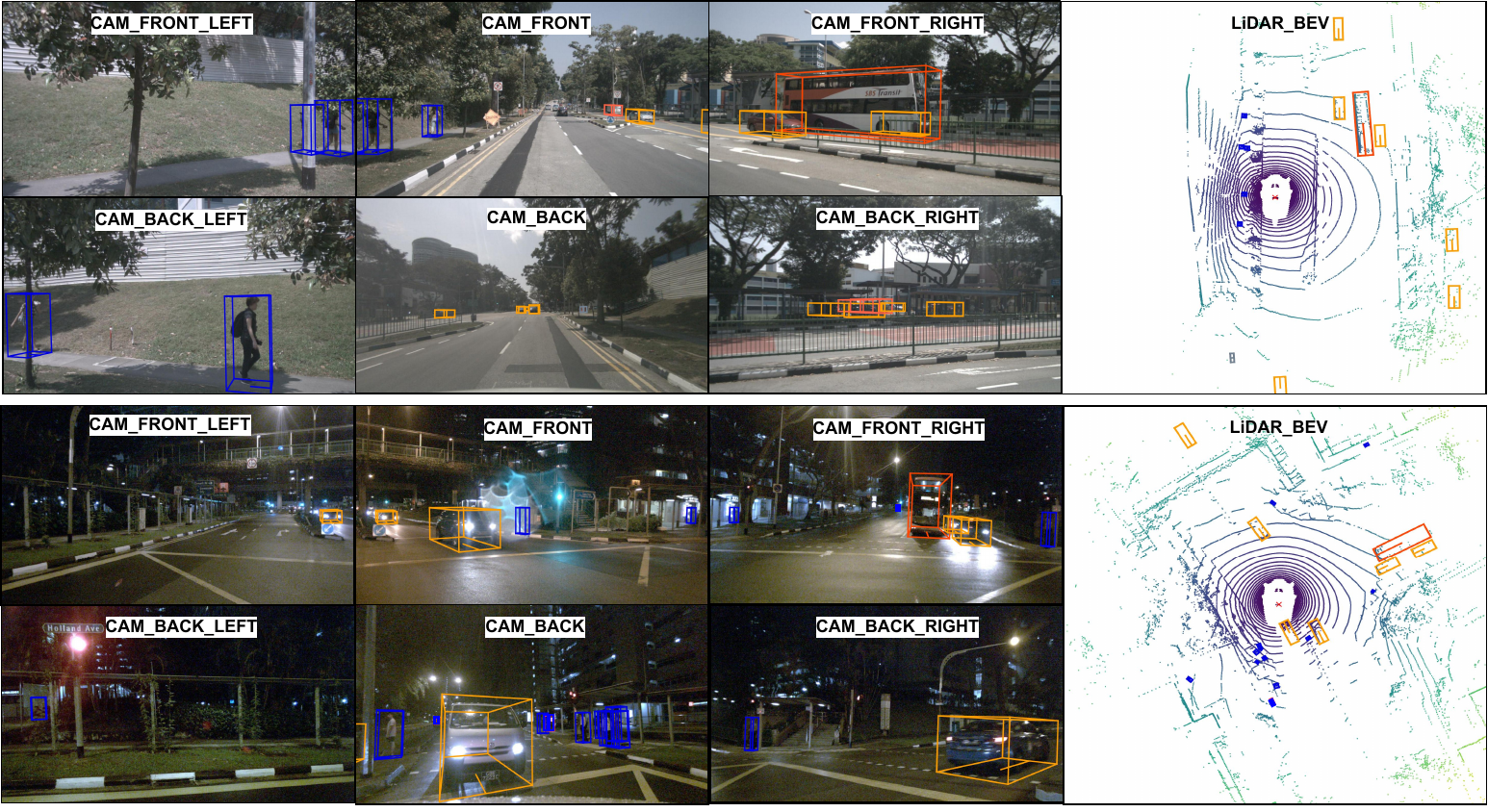}
    \captionsetup{font=small}
    \vspace{-0.1in}
    \caption{Qualitative visual results of our 3D object detection model on nuScenes\cite{caesar2020nuscenes} validation set (top) and test set (bottom) demonstrate its effectiveness in detecting distant, small, and occluded objects, including in nighttime scenes with bright light occlusion. Color-coded bounding boxes for \textcolor{blue}{pedestrians}, \textcolor{red}{buses}, and \textcolor{orange}{cars} highlight our model’s accuracy in distinguishing various object types.}

    
    \label{fig:result}
\end{figure*}

For evaluating catastrophic forgetting the metric we propose here is a bit different than what is mentioned
in \cite{aljundi2018memory}. Our metric involves initially training the model on
a fraction of labeled data in a supervised manner, followed
by evaluation on the test dataset. Subsequently, we train it
in the semi-supervised way with the remaining unlabeled
data using our proposed method. For evaluation, if $V_{1}$ and $V_{2}$ represent the number of objects correctly identified by
the supervised and the semi-supervised model respectively
when evaluated on the test data, then forgetting ($\Delta \%$) is defined as $\frac{{V_{1} - |V_{1} \cap V_{2}|}}{{V_{1}}} \times 100$, where $|V_{1} \cap V_{2}|$ denotes correctly
identified objects in both settings. Table \ref{tab:eval_data_percentage_nuscenes} shows that our Reflective-Teacher model best retains previously acquired knowledge. Fig. \ref{fig:result} presents qualitative results, highlighting the model’s ability to detect distant, small, and occluded objects, even in nighttime scenes with bright light occlusion.

\begin{table*}[!h]
    \centering
    \scalebox{.68}{
    \begin{tabular}{l|l|l|l|l|l|l|l|l|l|l|l|l|l
    }
    \hline
        Perspective & Modality & mAP & NDS & Car & Truck & CV & Bus & Trailer & Barrier & Motor & Bike & Ped & TC  \\ \hline
        $\times$     & LC  & 72.14 & 73.69 & 89.2  & 66.6  & 36.16 & 78.19 & 64.11 & 73.2 & 81.54 & 66.23 & 88.7 & 77.5  \\
        $\checkmark$ & LC  & 76.82  & 78.3  & 91.2  & 70.3  & 37.4  & 82.5  & 69.5  & 77.6 & 88.2  & 73.1  & 95.3 & 83.1  \\
        $\times$     & LCT & 75.29  & 77.14 & 87.17 & 68.11 & 44.64 & 75.96 & 69.1  & 80.3 & 82.7  & 66.1  & 92.6 & 86.22 \\
        $\checkmark$ & LCT & 79.31  & 80.4  & 92.7  & 71.5  & 47.1  & 77.9  & 70.4  & 84.6 & 88.5  & 73.6  & 95.2 & 91.6 \\ \hline
    \end{tabular}
    }
    \captionsetup{font=small}
    \vspace{-2mm}
     \caption{Evaluation on nuScenes test set \cite{caesar2020nuscenes} demonstrate the model's performance trained exclusively with $100 \%$ supervised data.}
     \label{tab:abl3}
\end{table*}

\begin{table*}[!h]
    \centering
    \scalebox{.68}{
     \begin{tabular}{l|l|l|l|l|l|l|l|l|l|l|l|l|l}
    \hline
        Perspective & Uncertainty & mAP & NDS & Car & Truck & CV & Bus & Trailer & Barrier & Motor & Bike & Ped & TC  \\ \hline
       $\times$     & $\times$     & 72.98      & 74.54 & 89.4  & 65.1  & 39.56 & 76.7  & 63.81 & 76.4                       & 81.12 & 66.75 & 87.32 & 83.73 \\
        $\times$     & $\checkmark$ & 75.92     & 76.77 & 90.7  & 68.34 & 42.65 & 78.52 & 67.76 & 76.69                      & 85.01 & 71.7  & 90.12 & 87.76 \\
        $\checkmark$ & $\times$ & 76.1 & 78.12 & 90.13 & 69.76 & 43.21 & 77.45 & 67.11 & 82. 65 & 86.13 & 70.59 & 92.17 & 88.34 \\
        $\checkmark$ & $\checkmark$ & 79.46 & 80.81  & 93.72  & 71    & 46.5  & 78.2  & 70.9  & 85.18 & 89.03 & 74.32  & 95.82 & 90.4 \\
        \hline
    \end{tabular}
    }
    \captionsetup{font=small}
    \vspace{-2mm}
     \caption{ Evaluation on nuScenes test set \cite{caesar2020nuscenes}  using LCT data, 
     demonstrating our model's performance trained with 25\% supervised data.  
    }
     \label{tab:abl5}
     \vspace{-3mm}
\end{table*}

\begin{table}[!ht]
    \centering
    \scalebox{.68}{
    \begin{tabular}{l|l|l|l|l}
    \hline
         Uncertainty &  5\% S & 10\% S & 20\% S & 25\% S \\ \hline
        $\times$ & 33.82 & 48.51 & 68.81 & 76.1 \\
        $\checkmark$ & 47.23 & 56.16 & 72.86 & 79.46 \\
         \hline
    \end{tabular}
    }
    \captionsetup{font=small}
    \caption{The improvement in mAP values on nuScenes test set with the incorporation of uncertainty into the model trained with different percentages of supervised data.}
     \label{tab:uncertainty_pct}
     \vspace{3.5mm}
\end{table}

\noindent
\textbf{On Waymo Open Dataset: }
We also conduct experiments on Waymo dataset \cite{sun2020scalability}. Our model's performance, evaluated using mean Average Precision (mAP) and mean Average Precision with Heading (mAPH) at two difficulty levels L1 and L2, shows superior results compared to previous methods. On the WOD validation set Table \ref{tab:waymo_val}, our model achieves (85.5\%, 83.2\%) in mAP and mAPH respectively for L1 and (78.6\%, 77.5\%) for L2 whereas for the test set, the scores are (85.9\%, 84.1\%) in mAP and mAPH for L1 and (78.4\%, 75.9\%) for L2.
Also, \textbf{Table \ref{tab:eval_data_percentage_nuscenes} shows that with 22\% of labeled data Waymo attends its equivalent supervised mAP value} as well as exhibits the greatest capability in retaining earlier knowledge.

\begin{figure}[!ht]
       \centering 
       \includegraphics[width=3.2in,height=1.7in]{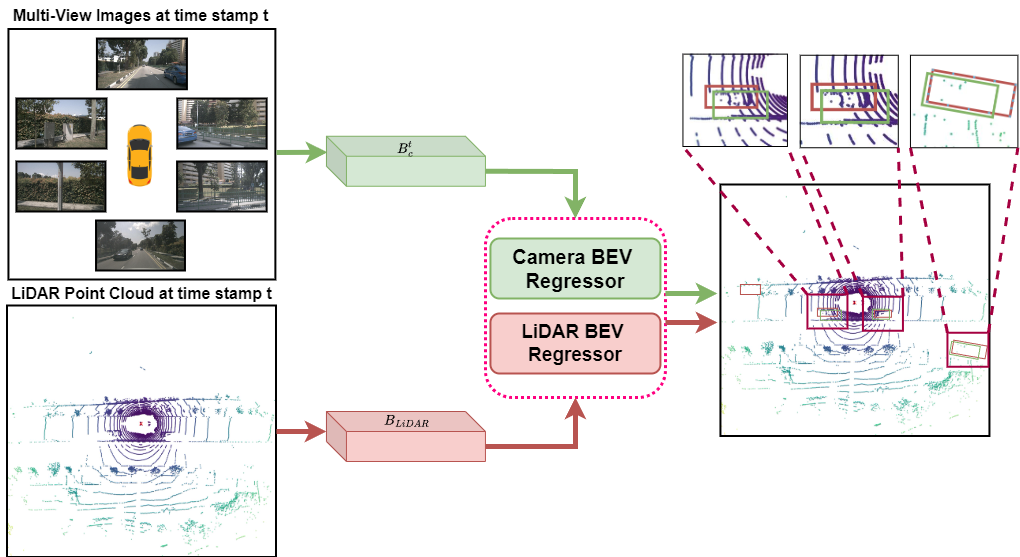}
        \captionsetup{font=small}
       \caption{Non-overlapping bounding boxes due to feature misalignment between LiDAR BEV and camera BEV.}
       \vspace{12pt}
       \label{fig:misalign}
\end{figure}

\subsection{Ablation study}

\textbf{Multimodal Feature Misalignment Analysis: }
To evaluate how our fusion model mitigates misalignment between different sensor modalities, we conducted a detailed analysis by integrating two separate regressor modules at the LiDAR BEV, $B^{t}_{LiDAR}$, and camera, $B^{t}_{c}$, during inference time. These regressors, which are integrated with pretrained object detection models, Bevformer V2 \cite{li2022bevformer} for camera sensor and VoxelNet \cite{zhou2018voxelnet} for LiDAR, have been trained on nuScenes dataset. As evident from Fig. \ref{fig:misalign}, there is a notable discrepancy between the bounding boxes predicted by the camera BEV regressor and the LiDAR BEV regressor, which diminishes significantly upon incorporating our proposed fusion module, making it robust for model deployment in the real world autonomous driving systems. Here we observe an IoU of 61.23\% on the nuScenes dataset and 67.74\% on the Waymo dataset.


\textbf{Incorporation of GA-BEVFusion:}
We perform an ablation study using a naive fusion approach where our proposed GA-BEVFusion is not used; instead, LiDAR BEV features and camera BEV features are concatenated elementwise and passed through a convolution module. We observe a performance drop compared to our fusion method. Our model without GA-BEVFusion achieves a score of $77.01\%$ mAP and $78.34\%$ NDS on the nuScenes test dataset, demonstrating a gain of $2.29\%$ mAP and $2.06\%$ NDS on inclusion of our proposed fusion strategy. In Fig. \ref{fig:misalign}, it is evident that normal fusion leads to misalignment between multimodal features, resulting in suboptimal bounding boxes. Hence, our fusion strategy proves to be beneficial for accurate 3D detection.

\textbf{Perspective supervision and temporal inputs:} Table \ref{tab:abl3} indicates a notable improvement in mAP and NDS values when considering perspective supervision in our model trained with the full labeled dataset for both scenarios, with and without temporal data. For LC (LiDAR and Camera) input data, the mAP and NDS improvements are 4.68\% and 4.61\%, respectively. For LCT (LiDAR, Camera, and Temporal) input data, the increases are 4.02\% in mAP and 3.26\% in NDS. This highlights that multi-view camera images and temporal information supervision improve our multimodal 3D detection performance.



\textbf{Incorporation of uncertainty measure:} We train our model on $25\%$ labeled data in our semi-supervised framework. The results are evaluated on the nuScenes test dataset and are listed in Table~\ref{tab:abl5}. As per earlier results, when training our model with $100\%$ supervised data, we observe mAP of $79.3\%$ and NDS of $80.4\%$. However, when it is trained using our semi-supervised approach with only $25\%$ labeled data, as shown in the third row of Table \ref{tab:abl5}, the mAP and NDS values are lower compared to when trained with $100\%$ supervised data. Upon incorporating uncertainty in our semi-supervised training approach, shown in the fourth row of Table \ref{tab:abl5}, we notice an incremental improvement of $0.2\%$ in mAP and $0.41\%$ in NDS, making it comparable with the supervised method. This indicates that the uncertainty measure helps the network in filtering out less confident regions, thereby enhancing overall accuracy. This is also evident from \textbf{Table \ref{tab:uncertainty_pct}, where mAP gradually increases with different proportions of supervised data and is higher than its without uncertainty counterpart.}

\textbf{Supervisory proportion:} Our ablation study examines mAP values across varying labeled data proportions. 
In Table \ref{tab:eval_data_percentage_nuscenes}, we compare our Reflective Teacher with other teacher-student based SSOD trained on our multimodal 3D object detection architecture. We notice all the semi-supervised methods greatly outperform the supervised method. Also we observe our Reflective Teacher consistently surpasses all other methods under different percentages of labeled data. Results show our model is able to achieve equivalent supervised performance on nuScenes dataset with only $25\%$ labeled data, emphasizing the efficiency of our semi-supervised approach.

In our work, we have considered only Unbiased Teacher \cite{liu2021unbiased} and Active Teacher \cite{mi2022active} for comparison in our ablation studies, as their source codes are available. Additionally, we conducted experiments using different multimodal models in our semi-supervised Reflective Teacher model, specifically BEVFusion \cite{liu2023bevfusion} and LIFT \cite{zeng2022lift}. Both models achieved equivalent accuracy with \textbf{more than 35\% labeled data on both nuScenes and Waymo dataset}, highlighting the effectiveness of our fusion approach.

%% file: include/conclusion.tex
\section{Conclusion}
In this paper, we propose Reflective Teacher, an efficient teacher-student multimodal learning approach for semi-supervised 3D object detection in BEV for autonomous driving, using LiDAR and camera data. Our proposed fusion strategy, GA-BEVFusion, then incorporating uncertainty measures and addressing catastrophic forgetting achieve state-of-the-art performance with just 25\% and 22\% of labeled data on nuScenes and Waymo dataset respectively. We demonstrate that the proposed end-to-end multimodal architecture is equally effective in fully supervised setup as well, thereby improving respective state-of-the-art approaches. 
